\documentclass[runningheads]{llncs}
\usepackage{graphicx}
\usepackage{comment}
\usepackage{amsmath,amssymb} 
\usepackage{color}

\usepackage{multirow}

\begin{document}
\pagestyle{headings}
\mainmatter
\def\ECCVSubNumber{1820}  

\title{Feedback Graph Convolutional Network for Skeleton-based Action Recognition} 

\titlerunning{FGCN for Skeleton-based Action Recognition}
%
\author{Hao Yang\inst{1} \and
Dan Yan\inst{1} \and
Li Zhang\inst{1,2} \and
Dong Li\inst{1,2} \and
YunDa Sun\inst{1} \and
Shaodi You\inst{3} \and
Stephen J. Maybank\inst{4}}
\authorrunning{H. Yang, D. Yan, L. Zhang, D. Li, Y. Sun, S. You, S. Maybank}
%
\institute{ R\&D Center of Artificial Intelligent, NUCTECH Company Limited, Beijing, China \email{\{yanghao1, yandan, sunyunda, li.dong\}@nuctech.com} \\
\and Department of Engineering Physics, Tsinghua University, Beijing, China \email{zli@mail.tsinghua.edu.cn} \\ 
\and Informatics Institute, University of Amsterdam. Amsterdam, Netherlands \email{s.you@uva.nl}
\and Department of Computer Science and Information Systems, Birkbeck College, London, United Kingdom \\
\email{sjmaybank@dcs.bbk.ac.uk}}

\maketitle

\begin{abstract}
	Skeleton-based action recognition has attracted considerable attention in computer vision since skeleton data is more robust to the dynamic circumstance and complicated background than other modalities. Recently, many researchers have used the Graph Convolutional Network (GCN) to model spatial-temporal features of skeleton sequences by an end-to-end optimization. However, conventional GCNs are feedforward networks which are impossible for low-level layers to access semantic information in the high-level layers. In this paper, we propose a novel network, named Feedback Graph Convolutional Network (FGCN). This is the first work that introduces the feedback mechanism into GCNs and action recognition. Compared with conventional GCNs, FGCN has the following advantages:
	(1) a multi-stage temporal sampling strategy is designed to extract spatial-temporal features for action recognition in a coarse-to-fine progressive process; (2) A dense connections based Feedback Graph Convolutional Block (FGCB) is proposed to introduce feedback connections into the GCNs. It transmits the high-level semantic features to the low-level layers and flows temporal information stage by stage to progressively model global spatial-temporal features for action recognition; (3) The FGCN model provides early predictions. In the early stages, the model receives partial information about actions. Naturally, its predictions are relatively coarse. The coarse predictions are treated as the prior to guide the feature learning of later stages for a accurate prediction.
	Extensive experiments on the datasets, NTU-RGB+D, NTU-RGB+D120 and Northwestern-UCLA, demonstrate that the proposed FGCN is effective for action recognition. It achieves the state-of-the-art performance on the three datasets.
	
\end{abstract}

\vspace{-4mm}
\section{Introduction}
\vspace{-1mm}

In recent years, the quantity of videos uploaded from various terminals has exploded. This has driven imperious demands for human action analysis automatically based on the content of videos. In particular, human action recognition using skeleton has attracted many computer vision researchers because of its strong adaptability to the effects of dynamic circumstance and complicated background, as compared with other modalities such as RGB \cite{karpathy2014large} and optical flow \cite{simonyan2014two}. Early deep learning methods using skeletons for action recognition usually represent the skeleton data as a sequence of joint-coordinate vectors \cite{du2015hierarchical,song2017end,zhang2017view,li2018independently} or a pseudo-image \cite{ke2017new,liu2017enhanced,liu2017two} which is then modeled by a RNN or CNN respectively. However, these methods do not explicitly exploit the spatial dependencies among correlated joints, even though the spatial dependencies are informative for understanding human actions. More recently, some methods \cite{yan2018spatial,shi2019two,shi2019skeleton,li2019actional} construct spatial temporal graphs based on the natural connections of joints and temporal edges of consecutive frames. They then exploit a GCN to model spatial-temporal features.
However, the conventional GCNs are all single-pass feedforward networks that are fed with the entire skeleton sequence. It is difficult for these methods to extract effective spatial-temporal features, because the useful information is usually buried in the motion-irrelevant or undiscriminating clips when they are fed with entire skeleton sequence. For example, in the action ``kicking something", most clips are ``standing upright", and in the action ``wear a shoe", most clips are a subject sitting on a chair.
Then the single-pass feedforward networks can not access the high-level semantic information for the low-level layers. Meanwhile, inputting the entire skeleton sequence increases computational complexity of the model.




Motivated by this, we propose a novel neural network, named Feedback Graph Convolutional Network (FGCN), to extract effective spatial-temporal features from skeleton data in a coarse-to-fine progressive process for action recognition. The FGCN is the first work that introduces feedback mechanism into GCNs and action recognition. Compared with conventional GCNs, the FGCN has a multi-stage temporal sampling strategy which divides input skeleton sequences into multiple stages in the temporal domain and sparsely samples input skeleton clips from temporal stages to avoid feeding with the entire skeleton sequence. Each sampled clip is input into graph convolutional layers to extract local spatial-temporal features for each stage. 
A Feedback Graph Convolutional Block (FGCB) is proposed to model global spatial-temporal features by fusing the local features. The FGCB is a local dense graph convolutional network with lateral connections from each stage to the next stage and it introduces feedback connections into conventional GCNs. From a semantic point of view it works in a top down manner, which makes it possible for low-level convolutional layers to access semantic information in the high-level layers at each stage. From the temporal domain, the feedback mechanism in FGCB works with a sequence of cause-and-effect and the output of the previous stage flows into the next stage to modulate its input.

Another advantage of the FGCN is that it provides early predictions of the output in a fraction of the total inference time. This is valuable in many applications such as robotics or autonomous driving, in which latency time is very crucial.
The early predictions are a result of the proposed multi-stage coarse-to-fine progressively optimization. In the early stages, FGCN is only fed with a part of skeleton sequence and the information about the action is limited, so the inferences of it are relatively coarse. These inferences are treated as a prior to guide the feature learning in later stages. In later stages, the model receives more complete information about the action and the guider of former inferences, thus it outputs more accurate inferences. Several temporal fusion strategies are proposed to fuse the local predictions in temporal stages for a video-level prediction. The strategies enable the network to be optimized in a progressive process.


The main contributions of this paper are summarized as follows:
\vspace{-1mm}
\begin{itemize}
	\item We propose a novel Feedback Graph Convolutional Network (FGCN) for action recognition from skeleton sequences. It models spatial-temporal features by a multi-stage progressive process. To our knowledge, this is the first work that introduces the feedback mechanism into GCNs and action recognition.
	\vspace{1mm}
	\item We propose a dense connections based Feedback Graph Convolutional Block (FGCB) which is a local network with lateral connections between two temporal stages. Functionally, it transmits high-level semantic features as priors to module its features in low-level layers.
	\vspace{1mm}
	\item The FGCN model provides early predictions, which benefits from the multi-stage coarse-to-fine progressive optimization. The proposed model is extensively evaluated on three datasets, NTU-RGB+D, NTU-RGB+D120 and Northwestern-UCLA, and it achieves state-of-the-art performance on the three datasets.
\end{itemize}

\vspace{-2mm}
\section{Related Works}
\vspace{-1mm}
\subsection{Skeleton based Action Recognition}

As the depth sensor technologies (\textit{i.e.} kinect \cite{zhang2012microsoft}) and pose estimation algorithms \cite{toshev2014deeppose,cao2017realtime} matured, it becomes possible to capture skeleton data in real time by locating the key joints. The skeleton data is robust to illumination change, scene variation, and complex background. These facilitate the data-driven method's development of skeleton-based action recognition.
Conventional action recognition methods usually extract hand-crafted features from skeleton sequences. Some traditional methods \cite{evangelidis2014skeletal,vemulapalli2014human,luo2013group,rahmani2015learning} design several view-invariant features of actions. Examples of these features are body part-based skeletal quads \cite{evangelidis2014skeletal,vemulapalli2014human}, group sparsity based class-specific dictionary coding \cite{luo2013group}, and canonical view transformed features \cite{rahmani2015learning}. Other traditional methods integrate the information from different modalities that are always available in 3D action datasets. Some works \cite{hu2015jointly,ohn2013joint,rahmani2014real,wang2013learning} combine the depth information with the skeleton to improve performance. The depth information is represented by HOG features \cite{hu2015jointly,ohn2013joint} and Fourier Temporal Pyramids \cite{wang2013learning}, or it is modeled by random decision forests \cite{rahmani2014real}.
The recent success of deep learning has led to a surge of deep network based skeleton modeling methods. The widely used models are RNNs and CNNs. RNN-based methods \cite{du2015hierarchical,song2017end,zhang2017view,li2018independently} usually concatenate all of the joint-coordinates (2D or 3D) in each frame as a vector and then model the features of actions by a RNN fed with a sequence of the coordinate vectors. CNN-based methods \cite{ke2017new,liu2017enhanced,liu2017two} stack the sequence of coordinate vectors to obtain a pseudo-image, and then reduce the action recognition using skeleton sequences to an image classification task. The two-stream based model \cite{zhao2017two} combines RNN and CNN, operating on coordinate vectors of skeletons and RGB images respectively, to improve performance from a single network. However, these methods do not explicitly model the spatial dependence between correlated joints which is crucial for understanding human actions.

\vspace{-1mm}
\subsection{GCN based Action Recognition}
The Graph Convolutional Networks (GCNs) \cite{bruna2014spectral,niepert2016learning,duvenaud2015convolutional,henaff2015deep,kipf2017semi} generalize the convolutional operation to deal with the data with graph construction. There are two main ways of constructing GCNs: spatial perspective and spectral perspective. Spatial perspective methods \cite{bruna2014spectral,niepert2016learning} directly perform the convolution filters on the graph vertexes and their neighbors. In contrast, spectral perspective methods \cite{duvenaud2015convolutional,henaff2015deep,kipf2017semi} consider the graph convolution as a form of spectral analysis by utilizing the eigenvalues and eigenvectors of the graph Laplacian matrices.
This work follows the spatial perspective based methods \cite{yan2018spatial,shi2019two,shi2019skeleton,li2019actional}. The ST-GCN model \cite{yan2018spatial} is proposed to move beyond the limitations of hand-crafted parts and traversal rules used in previous methods. It operates on a spatial temporal graph to model the structured information about the joints along both the spatial and temporal dimensions. Based on ST-GCN, the 2s-AGCN model \cite{shi2019two} proposes a two-stream adaptive graph convolutional network, which exploits the second-order information of the skeleton to improve the performance of action recognition. The DGNN model \cite{shi2019skeleton} represents the skeleton data as a directed acyclic graph based on the kinematic dependency between the joints and bones. The AS-GCN model \cite{li2019actional} proposes an actional-structural graph convolution network by generating the skeleton graph with actional links and structural links. However, conventional GCNs are all feedforward networks in which it is impossible for low-level layers to access the semantic information in high-level layers.

\vspace{-1mm}
\subsection{Feedback Network}

Feedback mechanism exists in the human visual cortex \cite{hupe1998cortical,gilbert2007brain}, and it has been a focus of research in psychology \cite{ashford1983feedback} and control theory \cite{lee1967foundations,parlos1994application}. In recent years, feedback mechanism has been introduced into deep neural networks in computer vision \cite{stollenga2014deep,zamir2017feedback,li2019feedback,haris2018deep,han2018image,carreira2016human}, because it allows the network to carry the information of output to correct previous states.
In object recognition, the dasNet model \cite{stollenga2014deep} exploits the feedback structure by dynamically altering its convolutional filter sensitivities during classification and iteratively focusing its internal attention on some of its convolutional filters. Feedback Network \cite{zamir2017feedback} firstly introduces the feedback mechanism into the convolutional recurrent neural network, which transfers the hidden state with high-level information to the input layer.
In super resolution, several efforts \cite{li2019feedback,haris2018deep,han2018image} are made to take advantage of the feedback mechanism. The DBPN model \cite{haris2018deep} proposes a deep back-projection network which exploits iterative up-projection and down-projection units to achieve error feedback. The DSRN model \cite{han2018image} proposes a dual-state RNN and transmits the information between two recurrent states via a delayed feedback. The SRFBN model \cite{li2019feedback} designs a feedback block to handle the feedback connections and refines low-level representations with high-level information.
In human pose estimation, \cite{carreira2016human} proposes an iterative error feedback (IEF) by iteratively estimating and applying a self-correction to the current estimation.

\vspace{-1mm}
\section{The Method}
\vspace{-1mm}
\subsection{Graph Convolutional Network}
\label{section_gcn}
GCNs generalize the convolution operation to learn effective representations from graph structured data. In action recognition, the skeleton of a body is defined as an undirected graph in which each joint of the skeleton is defined as a vertex of the graph and the natural connections in the human body are defined as edges of the graph. In this paper, the skeleton in the frame $t$ is denoted as a graph $G_t=\{\textbf{V}_t,\textbf{E}_t\}$, where $\textbf{V}_t$ is the set of joints in the frame and $\textbf{E}_t$ is the set of bones in the skeleton. For 3D skeleton data, the joint set is denoted as $\textbf{V}_t = \{v_{ti}\}^N_{i=1}$, where $v_{ti}=(x_{ti},y_{ti},z_{ti})$. Given two joints $v_{ti}=(x_{ti},y_{ti},z_{ti})$ and $v_{tj}=(x_{tj},y_{tj},z_{tj})$, a bone of the skeleton is defined as a vector $e_{v_{ti},v_{tj}}=(x_{tj}-x_{ti},y_{tj}-y_{ti},z_{tj}-z_{ti})$, $(i,j)\in Q$, where $Q$ is the set of naturally connected human body joints. The skeleton sequence with $len$ frames is denoted as $S= \{G_1, G_2,\dots, G_{len} \}$.

The graph convolution is defined operating on each vertex and its neighbors. For a vertex $v_{ti}$ in the graph, its neighbor set is denoted as $N(v_{ti}) =\{v_{tj}|d(v_{ti},v_{tj})\leq D\}$, where $d(v_{ti},v_{tj})$ is the length of the shortest path from $v_{tj}$ to $v_{ti}$. We set $D=1$ for the 1-distance neighbor set in this paper.
The graph convolution operating on the neighbor set of vertex $v_{ti}$ is formulated as:
\begin{equation}
\label{eqn_gconv}
f_{out}(v_{ti})=\sum_{v_{tj}\in N(v_{ti})} \frac{1}{Z[l(v_{tj})]} f_{in}(v_{tj}) W[l(v_{tj})],
\end{equation}
where $f_{in}$ and $f_{out}$ denote the input and output feature maps of this convolutional layer. $l(v_{tj})$ is the label function which allocates a label from $1$ to $K$ for the vertex in $N(v_{ti})$. In our experiments, we set $K = 3$ empirically to divide $N(v_{ti})$ into $3$ subsets. $W(\cdot)$ is the weighting function which provides a weight vector according to the label $l(v_{tj})$. Similarly, $Z[l(v_{tj})]$ denotes the number of vertexes corresponding to the subset of $l(v_{tj})$.

In implementation, the connections of a graph are recorded in an $N \times N$ adjacency matrix $\textbf{A}_k$. With the adjacency matrix, Eqn.~\ref{eqn_gconv} can be formulated as:
\begin{equation}
f_{out}=\sum_{k=1}^K \textbf{W}_k (\Lambda^{-\frac{1}{2}}_k \textbf{A}_k \Lambda^{-\frac{1}{2}}_k f_{in}) \odot (\textbf{M}_k),
\end{equation}
where $\odot$ denotes the dot product and $\Lambda^{ii}_k=\sum_{j} \textbf{A}^{ij}_k$ is a diagonal matrix. $\textbf{W}_k$ is the weight vector of the convolution operation, which corresponds to the weighting function $W(\cdot)$ in Eqn.~\ref{eqn_gconv}. In practice, $ \textbf{A}_k $ is allocated with a learnable weight matrix $\textbf{M}_k$ which is an $N\times N$ attention map that indicates the importance of each vertex. It is initialized as an all-one matrix.

\begin{figure}[t]
	\centering
	\includegraphics[width=0.95\linewidth]{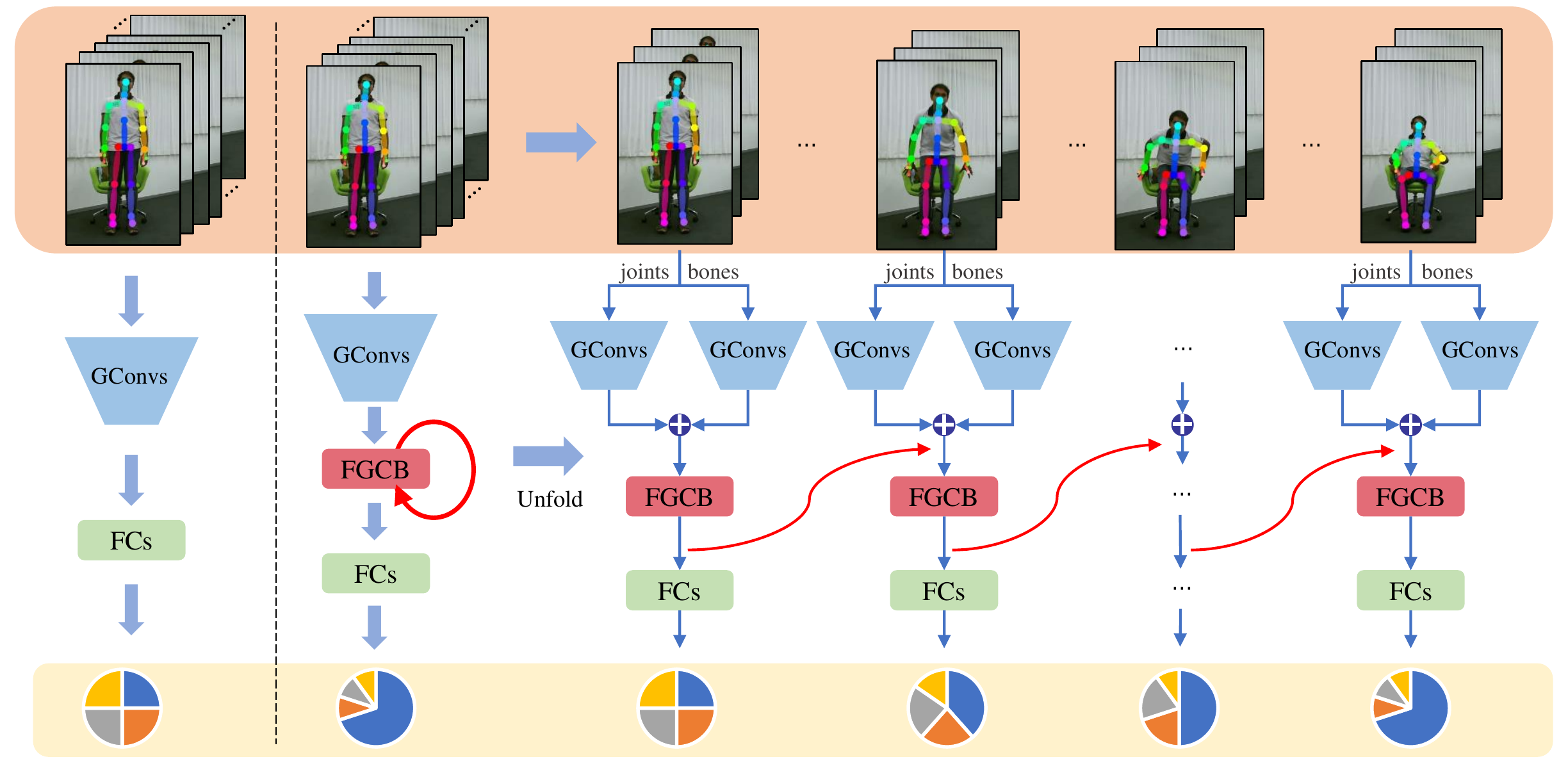}
	\caption{Comparison of the conventional GCNs (left) and the proposed FGCN (right). Red arrows represent the feedback connections of the feedback block (FGCB).}
	\vspace{-4mm}
	\label{fig-fgcn}
\end{figure}

\subsection{Feedback Graph Convolutional Network}
Traditional action recognition methods \cite{yan2018spatial,shi2019two,shi2019skeleton,li2019actional} based on GCNs are all fed with the entire skeleton sequence in a feedforward network. However, the useful information is usually buried in the motion-irrelevant and undiscriminating clips when fed with entire skeleton sequence. And single-pass feedforward networks can not access semantic information at low-level layers.
To tackle these problems, we propose a Feedback Graph Convolutional Network (FGCN) which extracts spatial-temporal features by a multi-stage progressive process, as shown in Fig.~\ref{fig-fgcn}. Specifically, in the FGCN a multi-stage temporal sampling strategy is designed to sparsely sample a sequence of input clips from the skeleton data, instead of operating on the entire skeleton sequence directly. These clips are first fed into graph convolutional layers to extract the local spatial-temporal features. Then, a Feedback Graph Convolutional Block (FGCB) is proposed to fuse the local spatial-temporal features from multiple temporal stages by transmitting the high-level information in the previous stage to the next stage to modulate its input. Finally, several temporal fusion strategies are proposed to fuse the local predictions from all temporal stages to give a video-level prediction.

Formally, given a skeleton sequence $S$, the multi-stage temporal sampling strategy first divides it into $T$ temporal stages with equal time interval, denoted as $S=\{s_1,s_2, \dots, s_T\}$. In each temporal stage, a skeleton clip is sampled randomly as an input of the deep model, denoted as $\{c_1,c_2, \dots, c_T\}$, where $c_t$ is the input clip sampled from the corresponding stage $s_t$. Each sampled clip $c_t$ is input to the stacked multiple graph convolutional layers to extract the local spatial-temporal features in the corresponding temporal stage, formulated as:
\begin{equation}
\textbf{F}_t=f_{GConvs}(c_t),
\end{equation}
where $t=1,2,\dots,T$, and $\textbf{F}_t$ is the local spatial-temporal features extracted by graph convolutional layers which are denoted as \textit{GConvs} in Fig.~\ref{fig-fgcn}.

\begin{figure}[t]
	\centering
	\includegraphics[width=0.95\linewidth]{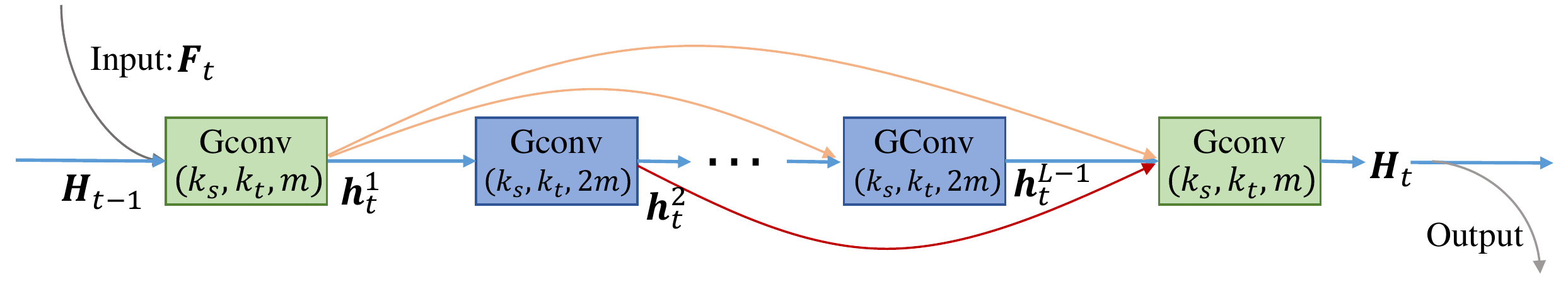}
	\caption{The detailed architecture of the proposed FGCB local network.}
	\vspace{-4mm}
	\label{fig-fgcb}
\end{figure}

The local features extracted from all temporal stages flow into the feedback block FGCB to learn global spatial-temporal features for action recognition. As shown in Fig.~\ref{fig-fgcb}, FGCB receives two inputs at the stage $t$: one is the hidden state from the previous stage $t-1$, denoted as $\textbf{H}_{t-1}$; the other is the local features from the current stage, denoted as $\textbf{F}_t$. Particularly, the input feature at the first stage $\textbf{F}_1$ is regarded as the initial hidden state $\textbf{H}_0$. Based on these two inputs, the feedback process of FGCB is formulated as:
\begin{equation}
\textbf{H}_t=f_{FGCB}(\textbf{H}_{t-1},\textbf{F}_t),
\end{equation}
where $\textbf{H}_t$ is the output of FGCB at stage $t$, and the function $f_{FGCB}(\cdot)$ represents the operations of the feedback block FGCB. More details about FGCB can be found in Section \ref{section fgcb}.

Following the FGCB, a fully connected layer and a softmax loss layer are used at each stage to predict actions. The prediction process from the output $H_t$ of FGCB is formulated as:
\begin{equation}
P_t = f_{pred}(\textbf{H}_t),
\end{equation}
where $P_t\in R^C$ denotes the local prediction at stage $t$ and $C$ is the number of actions. The function $f_{pred}(\cdot)$ represents the operations of the fully connected layer and the softmax layer. After operating on $T$ temporal stages, we will obtain totally $T$ local predictions $\{P_1,P_2,\dots,P_T\}$. Several temporal fusion strategies are proposed to fuse these local predictions corresponding to multiple stages for a video-level prediction $P_S$ which is computed as:
\begin{equation}
P_S = f_{tf}(P_1,P_2,\dots,P_T),
\end{equation}
where $f_{tf}$ is the operations of a temporal fusion strategy. In this paper, we propose three temporal fusion strategies, \textit{i.e.} last-win-all fusion, average fusion and weighting fusion. The FGCN model is trained end-to-end with the cross-entropy loss as follows:
\begin{equation}
L(y,P_S) = -\sum_{i=1}^{C}y^i \log(P_S^i),
\end{equation}
where $y$ is the action label of the skeleton $S$, if $y=i$, $y^i$ is set as $1$, otherwise it is set as $0$.

\subsection{Feedback Graph Convolutional Block}
\label{section fgcb}

The feedback block FGCB is the core component of the FGCN model. On the one hand, the FGCB transmits the high-level semantic information back to low-level layers to refine their encoded features. On the other hand, the output at the previous stage flows into the next stage to modulate its input. To enable the FGCB to effectively transmit information from high-level to low-level and from the previous stage to the next stage, we propose a dense connected local graph convolutional network which adds shortcut connections from each layer to all subsequent layers. At a temporal stage $t$, the FGCB receives the high-level information from the output $\textbf{H}_{t-1}$ of the previous stage to modulate the low-level input $\textbf{F}_t$ of the current stage. In our model, the FGCB consists of $L$ spatial temporal graph convolutional layers. The spatial temporal graph convolutional layer is denoted as $GConv(k_s,k_t,m)$ in Fig.~\ref{fig-fgcb}, where $k_s$ and $k_t$ are the kernel size in the spatial and temporal domains respectively, and $m$ denotes output channels of the graph convolutional layer.

As shown in Fig.~\ref{fig-fgcb}, the first convolutional layer in FGCB receives two inputs $\textbf{F}_t$ and $\textbf{H}_{t-1}$. It compresses and fuses the features from the concatenation of the two inputs $[\textbf{F}_t,\textbf{H}_{t-1}]$. The output of this layer is formulated as:
\begin{equation}
\textbf{h}_t^1 = f_{FGCB}^1([\textbf{F}_t,\textbf{H}_{t-1}]),
\end{equation}
where $f_{FGCB}^1(\cdot)$ denotes the operations in the first convolutional graph layer of FGCB, and $\textbf{h}_t^1$ denotes the output feature maps of the first layer. Following the first layer, the $l_{th}$ layer receives the output feature maps from all preceding layers, $\textbf{h}_t^1, \textbf{h}_t^2,\dots,\textbf{h}_t^{l-1}$, as input:
\begin{equation}
\textbf{h}_t^l=f_{FGCB}^l([\textbf{h}_t^1, \textbf{h}_t^2,\dots,\textbf{h}_t^{l-1}]),
\end{equation}
where $l=1,2,\dots,L$ and $ [\textbf{h}_t^1, \textbf{h}_t^2,\dots,\textbf{h}_t^{l-1}]$ refers to the concatenated feature maps in preceding layers. Similar to the first layer, the final layer in FGCB compresses and fuses the feature maps from the outputs of all preceding layers to produce the output of FGCB:
\begin{equation}
\textbf{H}_t=\textbf{h}_t^L=f_{FGCB}^L([\textbf{h}_t^1, \textbf{h}_t^2,\dots,\textbf{h}_t^{L-1}]),
\end{equation}
\vspace{-6mm}

\begin{figure}[t]
	\centering
	\includegraphics[width=0.9\linewidth]{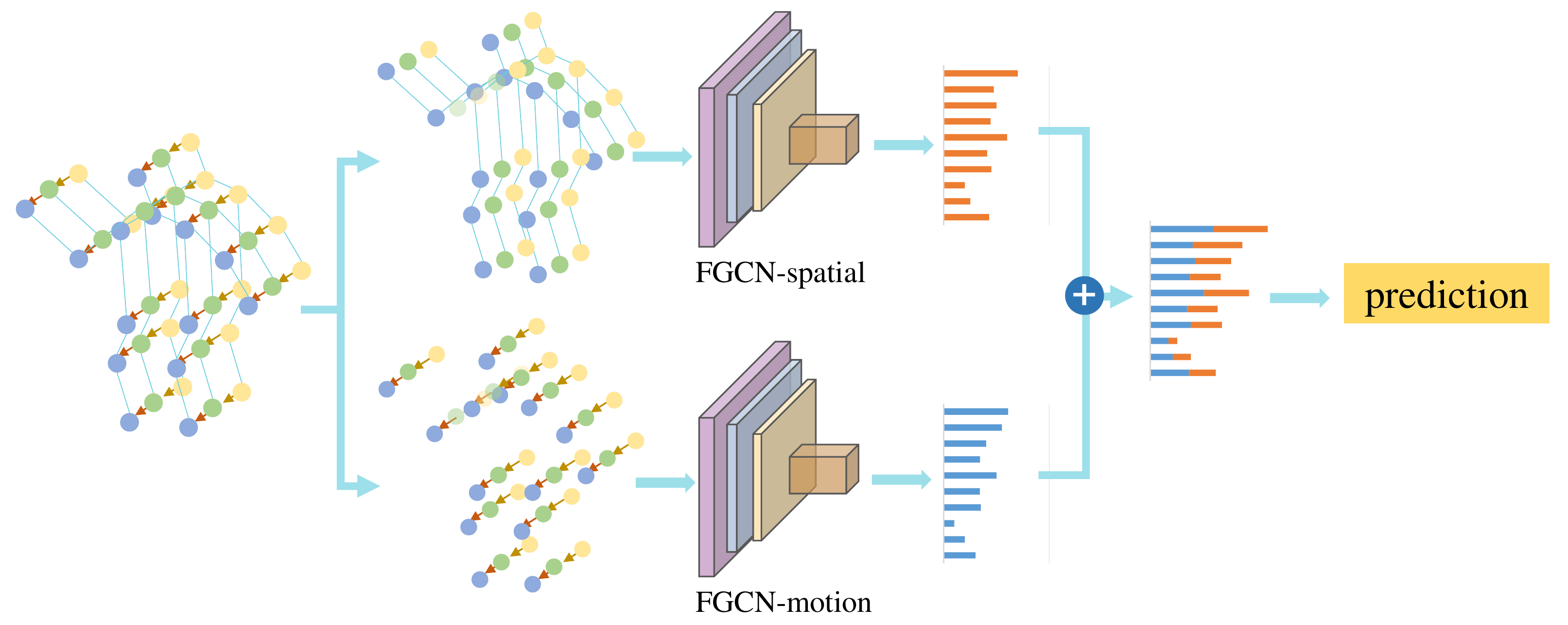}
	\caption{The prediction scores of FGCN-spatial and FGCN-motion are fused for final action prediction.}
	\vspace{-4mm}
	\label{fig-2stream}
\end{figure}

\subsection{Two-stream Framework of FGCN}
The joints and bones of a skeleton only contain spatial information of actions, However, many actions are difficult to recognize from the spatial information alone, for example ``wear a shoe" versus ``take off a shoe", ``wear on glasses" versus ``take off glasses" and \textit{etc}. Inspired by \cite{shi2019skeleton}, we model the spatial-temporal features not only exploiting spatial information but the temporal movement information of skeleton sequences. As in Section \ref{section_gcn}, the joint and bone of skeleton are denoted as a vector of coordinates. The movement of a joint or bone is defined as the difference of the vectors for the same joint or bone in consecutive frames along the temporal dimension.

Given the joints and bones from two consecutive frames, denoted as $v_{ti}$, $v_{(t+1)i}$ and $e_{v_{ti},v_{tj}}$, $e_{v_{(t+1)i},v_{(t+1)j}}$ respectively, the movement of joints is defined as $mv_{ti}=v_{(t+1)i}-v_{ti}$. Similarly, the movement of bones is defined as $me_t^{ij}=e_{v_{(t+1)i},v_{(t+1)j}}-e_{v_{ti},v_{tj}}$.
As the spatial information modeling, the motion information is formulated as a sequence of graphs 
$S^m = \{G^m_1, G^m_2,\dots, G^m_{len} \}$, where $G^m_t=\{\textbf{V}^m_t,\textbf{E}^m_t\}$, $\textbf{V}^m_t=\{mv_{ti}\}_{i=1}^N$ and $\textbf{E}^m_t=\{me_t^{ij}\}_{(i,j)\in Q}$. In this paper, the spatial graph $S$ and the motion graph $S^m$ are fed into two separate FGCN models to predict action labels. The model fed with spatial graphs $S$ is denoted as FGCN-spatial, the other fed with temporal graphs $S^m$ is denoted as FGCN-motion. The two models are finally fused by weighting the output scores of the softmax layers, as shown in Fig.~\ref{fig-2stream}.

\vspace{-1mm}
\section{Experiments}
\vspace{-1mm}
In this section, we evaluate the proposed FGCN method by conducting extensive experiments on three 3D skeleton action datasets, NTU-RGB+D, NTU-RGB+D120, and Northwestern-UCLA.

\vspace{-1mm}
\subsection{Datasets}

\hspace{5mm}\textbf{NTU-RGB+D} \cite{shahroudy2016ntu} is a widely used dataset for skeleton-based action recognition. The dataset contains more than 56,000 skeleton sequences categorized into 60 action classes. It provides 25 major body joints with 3D coordinates for every human in each frame. Two benchmark evaluations are recommended: cross-subject and cross-view. For cross-subject, both training and test sets consist of 20 subjects, and have 40,320 and 16,560 sequences respectively. The cross-view setup divides the data according to camera views. The training set has 37,920 sequences captured from the front and two side views, while the test set has 18,960 sequences captured from left and right 45 degree views.

\textbf{NTU-RGB+D120} \cite{liu2019ntu} is currently the largest in-door captured 3D skeleton dataset. It is an extension of NTU-RGB+D with 120 action classes and more than 114,000 video samples. The newly added action classes make the action recognition more challenging. For example, different actions may have similar body motions but different subjects. There may be fine-grained hand or finger motions and so on. The dataset has 106 subjects and 32 setup IDs. Cross-subject and cross-setup benchmarks are defined. For cross-subject, 53 subjects constitute the training set, and the remaining 53 subjects constitute the test set. Analogously, the 32 setup IDs are also divided equally into two parts for training and testing in cross-setup.


\textbf{Northwestern-UCLA} \cite{wang2014cross} is a multi-view 3D event dataset captured simultaneously by three Kinect cameras from different viewpoints. This dataset includes 1494 video sequences covering 10 action categories performed by 10 subjects from 1 to 6 times. It provides 3D spatial coordinates of 20 major body joints. As reported in \cite{wang2014cross}, we pick all samples from the first two cameras for training. The samples from the remaining cameras are for testing.

\begin{figure}[t]
	\centering
	\includegraphics[width=0.9\linewidth]{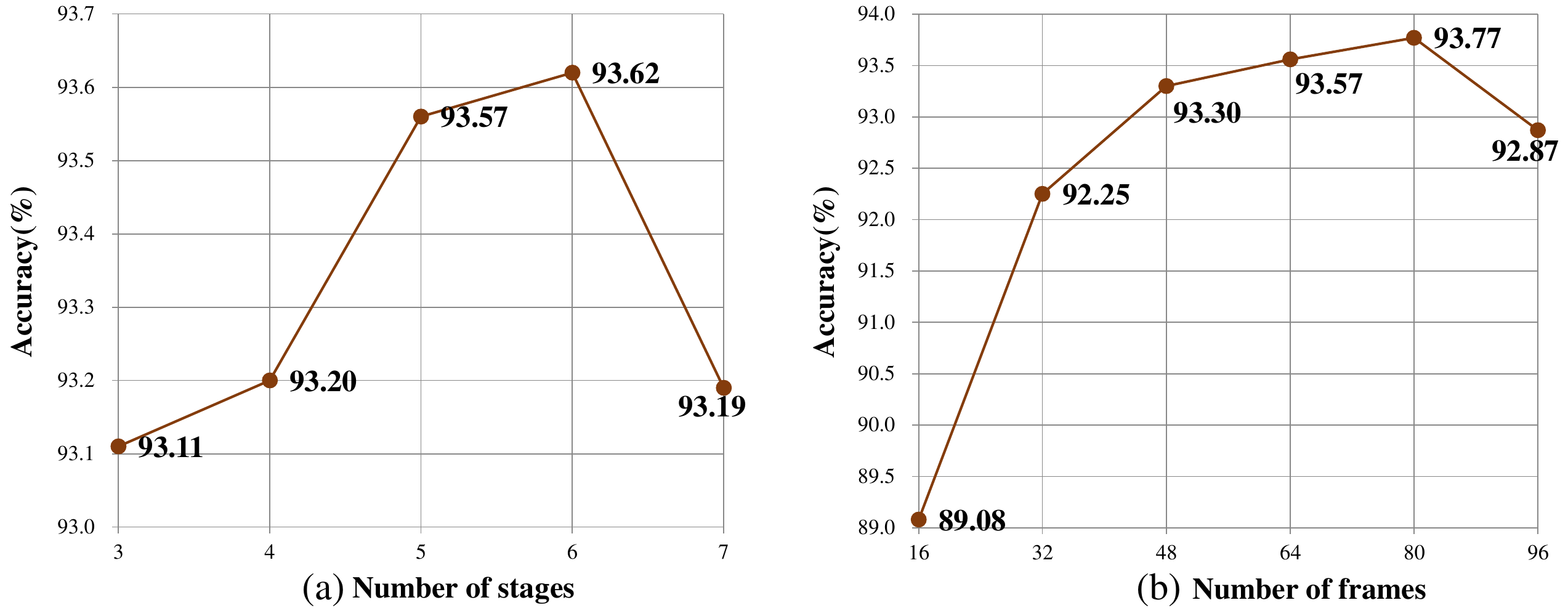}
	\caption{Evaluating the influence of two key factors on NTU-RGB+D, (a) influence of the number of stages, (b) influence of frame length in each stage.}
	\vspace{-4mm}
	\label{fig-numf-nums}
\end{figure}

\vspace{-1mm}
\subsection{Implementation Details}

All experiments are implemented with PyTorch deep learning framework. A stochastic gradient descent (SGD) optimizer is used during training with the batch size as 32, the momentum as 0.9, and the initial learning rate as 0.1. The learning rate is divided by 10 at the 40th and 60th epoch. The training process ends at the 80th epoch. In our experiments, the input video is divided into five stages temporally and 64 consecutive frames are sampled randomly from each stage to form an input clip.
Ten graph convolutional layers are stacked at the front of the feedback block FGCB and these layers have the same configuration as the graph convolutional layers in ST-GCN \cite{yan2018spatial}. The FGCB has four graph convolutional layers (\textit{i.e.} $L=4$). The spatial temporal kernel sizes and output channels of them are set as $k_s=3$, $k_t=3$ and $m=256$ respectively.

\begin{table}[ht]
	\caption{ Evaluating different temporal fusion strategies on NTU-RGB+D.}
	\label{tab-temporal-fusion}
	\tabcolsep=6pt
	\centering
	\begin{tabular}{ccccccc}
		\hline
		\multirow{2}{*}{Temporal Fusion Strategies} &  \multicolumn{5}{c}{ Weights } &\multirow{2}{*}{Cross-view(\%)} \\
		\cline{2-6}
		& $w_1$ & $w_2$ & $w_3$ & $w_4$ &  $w_5$ & \\
		\hline		
		Last-win-all fusion &0&0&0&0&1&89.88 \\
		Weight fusion-1 &0.05&0.05&0.1&0.2&0.6&93.09 \\
		Weight fusion-2 &0.1&0.15&0.2&0.25&0.3&93.05 \\
		Average fusion &0.2&0.2&0.2&0.2&0.2&\textbf{93.57} \\
		\hline
	\end{tabular}
\end{table}

\vspace{-1mm}
\subsection{Ablation Study}
In this section, we design four ablation experiments to evaluate the influence of different hyper-parameters, architecture and inputs on the performance of our FGCN model. These ablation experiments are all conducted on the challenging skeleton dataset NTU-RGB+D.

In the first experiment, we evaluate the influence of two key hyper-parameters on the performance of our FGCN model, \textit{i.e.}, the number of stages and the length of the input clip in each stage. In Fig.~\ref{fig-numf-nums}(a), the performances of FGCN with different numbers of temporal stages are reported. The FGCN model achieves the best performance when the input video is divided into 6 stages with equal duration. In the subsequent experiments, we set the number of temporal stages at 5, to balance performance against computational cost. Similar performances are obtained with 6 temporal stages. In Fig.~\ref{fig-numf-nums}(b), we evaluate the performance of FGCN fed with different numbers of frames at each stage. Based on the similar model selection strategy in the last experiment, we set the frame length as 64 in the subsequent experiments to balance performance against computational cost.

\begin{figure}[ht]
	\centering
	\includegraphics[width=0.95\linewidth]{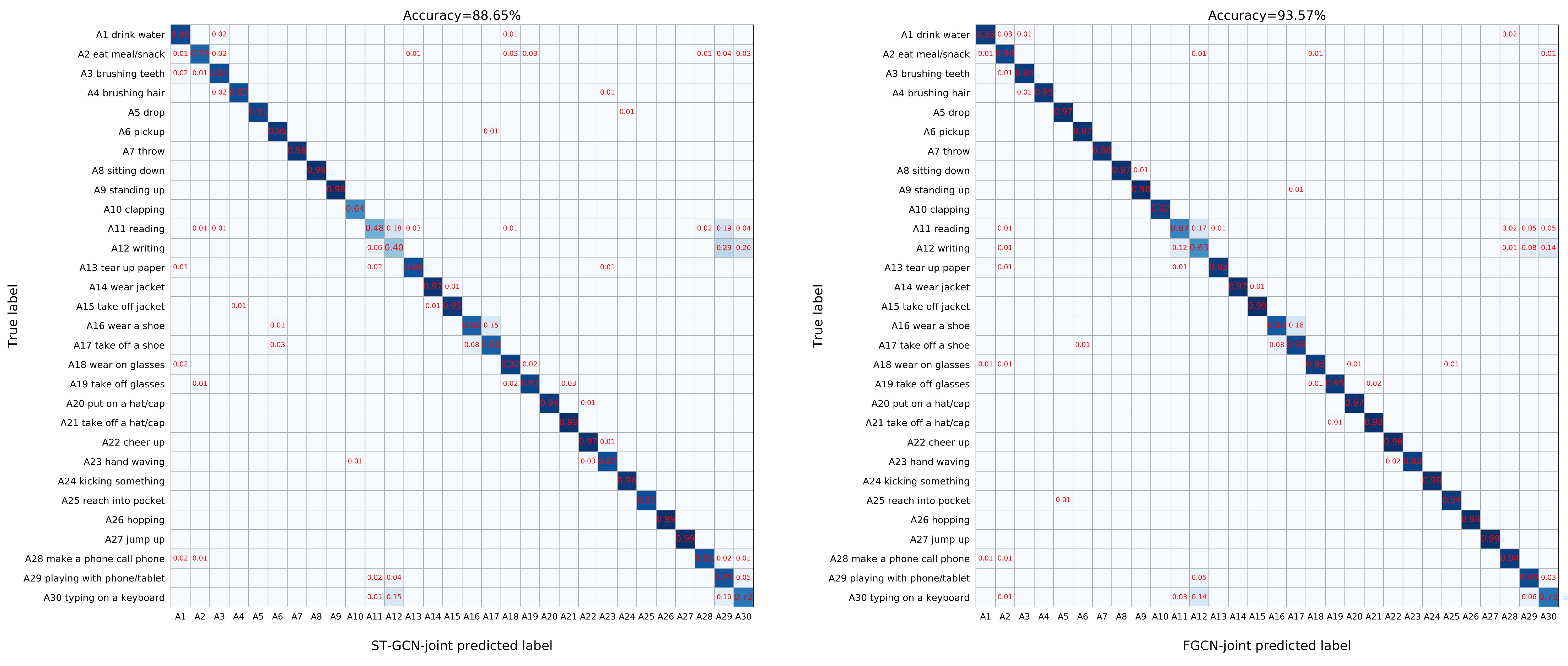}
	\caption{The confusion matrices of ST-GCN-joint and FGCN-joint on NTU-RGB+D.}
	\vspace{-4mm}
	\label{fig-confusion}
\end{figure}

In the second experiment, we evaluate the effectiveness of different temporal fusion strategies in the FGCN model, \textit{i.e.} last-win-all fusion, average fusion, and weight fusion. The experiment results are listed in Tab.~\ref{tab-temporal-fusion}. Among these three fusion strategies, the average fusion strategy achieves the best performance. Based on the results, we use the average fusion strategy to fuse the local predictions for the video-level prediction in the subsequent experiments.

\begin{table}[t]
	\caption{Evaluating the effectiveness of the FGCN model fed with different inputs on the NTU-RGB+D dataset.}
	\label{tab-joint-bone}
	\tabcolsep=6pt
	\centering
	\begin{tabular}{ccc}
		\hline
		Models & Cross-subject(\%) & Cross-view(\%) \\
		\hline
		ST-GCN-joint \cite{yan2018spatial} &81.5 &88.3\\
		FGCN-joint &87.04 &93.57	\\
		FGCN-bone	&86.96 &93.22 \\
		FGCN-joint+FGCN-bone	&89.24 &95.28 \\	
		\hline
		FGCN-spatial &88.32 &94.82	\\
		FGCN-motion	&85.96 &93.57 \\
		FGCN-spatial+FGCN-motion	&\textbf{90.22} &\textbf{96.25} \\	
		\hline
	\end{tabular}
\end{table}

In the third experiment, we evaluate the effectiveness of the proposed FGCN model fed with joints and bones. We first compare the proposed FGCN model with its baseline ST-GCN model. The two models have the same architecture and configuration of convolutional layers. As shown in the upper part of Tab.~\ref{tab-joint-bone}, the FGCN model fed with joint sequences of skeletons (FGCN-joint) outperforms the baseline model ST-GCN-joint by 5.54\% and 5.27\% on the cross-subject and cross-view benchmarks respectively. The confusion matrices for the former 30 classes are shown in Fig.~\ref{fig-confusion}, and the complete confusion matrices are shown in the supplementary materials. The improvements indicate that introducing feedback mechanism into GCNs is very effective for action recognition. Moreover, we fuse the softmax scores of two FGCN models, where one model is FGCN-joint, the other is FGCN-bone which is fed with the bone sequences. The fusion model FGCN-joint+FGCN-bone achieves a clear improvement, compared with FGCN-joint and FGCN-bone.


In the fourth experiment, we evaluate the effectiveness of FGCN model fed with the spatial information and the motion information, \textit{i.e.} FGCN-spatial and FGCN-motion, on the NTU-RGB+D dataset. The experiment results of these two models and their fusion are reported in the under part of Tab.~\ref{tab-joint-bone}. Firstly, the FGCN-spatial model fed with spatial information (joints and bones) achieves 88.32\% on cross-subject and 94.82\% on cross-view. It is comparable with the performance of the FGCN-joint+FGCN-bone model that fuses the softmax scores of two models. Then, the FGCN-motion fed with the movement of joints and bones achieves 85.96\% on cross-subject and 93.57\% on cross-view. Finally, we fuse the softmax scores of FGCN-spatial and FGCN-motion. The FGCN-spatial+FGCN-motion achieves 90.22\% on cross-subject and 96.25\% on cross-view, and it achieves a clear improvement from both of FGCN-spatial and FGCN-motion.

\begin{table}[t]
	\caption{Comparisons with the state-of-the-art methods on NTU-RGB+D.}
	\label{tab-state-rgbd}
	\tabcolsep=6pt
	\centering
	\begin{tabular}{ccc}
		\hline
		Models & Cross-subject(\%) & Cross-view(\%) \\
		\hline		
		ResNet152-3S (ICMEW 2017) \cite{li2017skeleton}	&85.0	&92.3 \\
		ST-GCN (AAAI 2018) \cite{yan2018spatial}	&81.5	&88.3 \\
		DPRL+GCNN (CVPR 2018) \cite{tang2018deep}	&83.5	&89.8 \\
		SR-TSL (ECCV 2018) \cite{si2018skeleton}	&84.8	&92.4 \\
		PB-GCN (BMVC 2018) \cite{thakkar2018part}	&87.5	&93.2 \\
		Bayesian GC-LSTM (ICCV 2019) \cite{zhao2019bayesian}	&81.8	&89.0 \\
		AS-GCN (CVPR 2019) \cite{li2019actional}	&86.8	&94.2 \\
		AGC-LSTM (CVPR 2019) \cite{si2019attention}	&89.2	&95.0 \\
		2s-AGCN (CVPR 2019) \cite{shi2019two}	&88.5	&95.1 \\
		DGNN (CVPR 2019) \cite{shi2019skeleton}	&89.9	&96.1 \\	
		\hline
		FGCN (ours)	&\textbf{90.2}	&\textbf{96.3} \\
		\hline
	\end{tabular}
\end{table}

\vspace{-1mm}
\subsection{Comparison with State-of-the-art}

In this section, we compare the performance of the FGCN model with the recent state-of-the-art methods on the NTU-RGB+D dataset, the NTU-RGB+D120 dataset, and the Northwestern-UCLA dataset.

For the NTU-RGB+D dataset, we display the accuracy of skeleton based action recognition methods, such as CNN-based methods \cite{li2017skeleton}, RNN-based methods \cite{si2018skeleton,si2019attention,zhao2019bayesian} and GCN based methods \cite{li2019actional,shi2019skeleton,shi2019two,yan2018spatial}. As shown in Tab.~\ref{tab-state-rgbd}, the proposed FGCN model achieves 8.7\% and 8.0\% improvements on the cross-subject and cross-view benchmarks respectively over the most comparable method ST-GCN \cite{yan2018spatial}. These improvements show the effectiveness of the proposed feedback framework in action recognition. Moreover, the FGCN model outperforms other recent state-of-the-art methods, such as AS-GCN \cite{si2019attention}, 2s-AGCN \cite{shi2019two}, and DGNN \cite{shi2019skeleton}. Our FGCN model achieves state-of-the-art performance on both cross-subject and cross-view benchmarks of the NTU-RGB+D dataset.

\begin{table}[b]
	\caption{Comparisons with the state-of-the-art methods on NTU-RGB+D120.}
	\label{tab-state-rgbd120}
	\tabcolsep=4pt
	\centering
	\begin{tabular}{ccc}
		\hline
		Models & Cross-subject(\%) & Cross-setup(\%) \\
		\hline		
		Internal Feature Fusion (T-PAMI 2017) \cite{liu2017skeleton}	&58.2	&60.9 \\
		Multi-Task Learning Network (CVPR 2017) \cite{ke2017new}	&58.4	&57.9 \\
		Skeleton Visualization (PR 2017) \cite{liu2017enhanced}	&60.3	&63.2\\
		Two-Stream Attention LSTM (TIP 2017) \cite{liu2017skeleton-tip} 	&61.2	&63.3 \\
		Multi-Task CNN with RotClips (TIP 2018) \cite{ke2018learning}	&62.2	&61.8 \\
		ST-GCN (AAAI 2018) (reported in \cite{papadopoulos2019vertex})	&72.4 & 71.3	\\
		AS-GCN (CVPR 2019) (reported in \cite{papadopoulos2019vertex}) &77.7 & 78.9 \\
		FSNet (T-PAMI 2019) \cite{liu2019skeleton}	&59.9	&62.4 \\
		TSRJI (SIBGRAPI 2019) \cite{caetano2019skeleton} &67.9	&62.8  \\
		LSTM-IRN (arXiv 2019) \cite{perez2019interaction}	&77.7	&79.6  \\
		GVFE + AS-GCN (arXiv 2019) \cite{papadopoulos2019vertex}	&78.3	&79.8 \\
		\hline
		FGCN (ours)	&\textbf{85.4}	&\textbf{87.4} \\
		\hline
	\end{tabular}
\end{table}

For the UTU-RGB+D120 dataset, the results on cross-subject and cross-setup benchmarks of the recent state-of-the-art methods are listed in Tab.~\ref{tab-state-rgbd120}. The proposed FGCN model achieves 85.4\% on cross-subject and 87.4\% on cross-setup and it outperforms the most comparable ST-GCN model \cite{yan2018spatial} by 13.0\% and 16.1\% on the cross-subject and cross-setup benchmarks respectively. The FGCN model outperforms other state-of-the-art methods with much lager margins. For example, the FGCN model outperforms Two-Stream Attention LSTM \cite{liu2017skeleton-tip} by over 24\% on both cross-subject and cross-setup benchmarks, and outperforms the most recent work FSNet \cite{liu2019skeleton} by over 25\% on both of cross-subject and cross-setup benchmarks.

For the typical 3D action recognition dataset Northwestern-UCLA, we compare the proposed FGCN model with the state-of-the-art methods in recent years. The results of these models are reported in Tab.~\ref{tab-state-ucla}. The FGCN model outperforms the part-based hierarchical recurrent neural network HBRNN-L \cite{du2015hierarchical} by 16.8\%. The recent method AGC-LSTM proposes an attention enhanced graph convolutional LSTM network to capture discriminative features from the co-occurrence relationship between spatial configuration and temporal dynamics. The FGCN model outperforms it by 2\%. Moreover, the FGCN model outperforms the most recent methods, such as JS+JM+BS+BM \cite{li2019learning}, HiGCN \cite{huang2019hierarchical} and MSNN \cite{shao2020learning}. The proposed FGCN model achieves state-of-the-art performance on the Northwestern-UCLA dataset.

\begin{table}[t]
	\caption{Comparisons with the state-of-the-art methods on Northwestern-UCLA.}
	\label{tab-state-ucla}
	\tabcolsep=10pt
	\centering
	\begin{tabular}{cc}
		\hline
		Models & Accuracy(\%) \\
		\hline
		Actionlet ensemble (T-PAMI 2013) \cite{wang2013learning}	&76.0 \\	
		Lie group (CVPR 2014 ) \cite{vemulapalli2014human}	&74.2 \\
		HBRNN-L(CVPR 2015) \cite{du2015hierarchical}	&78.5 \\
		Skeleton Visualization (PR 2017) \cite{liu2017enhanced}	&86.1 \\
		Ensemble TS-LSTM (ICCV 2017) \cite{lee2017ensemble}	&89.2 \\
		AGC-LSTM (CVPR 2019) \cite{si2019attention}	&93.3 \\
		JS+JM+BS+BM (ICME 2019) \cite{li2019learning} & 91.3 \\
		HiGCN (ICIG 2019) \cite{huang2019hierarchical} & 88.9 \\
		MSNN (CSVT 2020) \cite{shao2020learning} & 89.4 \\
		\hline
		FGCN (ours)	&\textbf{95.3} \\
		\hline
	\end{tabular}
\end{table}

\vspace{-1mm}
\section{Conclusion}
In this paper, we propose a novel FGCN model to extract effective spatial-temporal features of actions in a coarse-to-fine progressive process. Firstly, we propose a multi-stage temporal sampling strategy to sample sparse skeleton clips in multiple temporal stages and exploit graph convolutional layers to extract local spatial-temporal features for each stage. Then, we introduce the feedback mechanism into conventional GCNs by proposing the FGCB which is a local graph convolutional dense network. The FGCB transmits the semantic information from high-level layers to low-level layers and from the former stages to the later stages. Moreover, the FGCN provides early predictions which help agents in many applications to make timely decisions on-the-fly.
The proposed FGCN model is extensively evaluated on the NTU-RGB+D, NTU-RGB+D120 and Northwestern-UCLA datasets, indicating that the FGCN is effective for action recognition. It has achieved state-of-the-art performance on the three datasets.

%
%
\bibliographystyle{splncs04}
\bibliography{egbib}

\begin{thebibliography}{10}
\providecommand{\url}[1]{\texttt{#1}}
\providecommand{\urlprefix}{URL }
\providecommand{\doi}[1]{https://doi.org/#1}

\bibitem{ashford1983feedback}
Ashford, S.J., Cummings, L.L.: Feedback as an individual resource: Personal
  strategies of creating information. Organizational Behavior and Human
  Performance  \textbf{32}(3),  370--398 (1983)

\bibitem{bruna2014spectral}
Bruna, J., Zaremba, W., Szlam, A., Lecun, Y.: Spectral networks and locally
  connected networks on graphs. In: International Conference on Learning
  Representations (2014)

\bibitem{caetano2019skeleton}
Caetano, C., Br{\'e}mond, F., Schwartz, W.R.: Skeleton image representation for
  {3D} action recognition based on tree structure and reference joints. In:
  SIBGRAPI Conference on Graphics, Patterns and Images. pp. 16--23. IEEE (2019)

\bibitem{cao2017realtime}
Cao, Z., Simon, T., Wei, S.E., Sheikh, Y.: Realtime multi-person {2D} pose
  estimation using part affinity fields. In: IEEE Conference on Computer Vision
  and Pattern Recognition. pp. 7291--7299 (2017)

\bibitem{carreira2016human}
Carreira, J., Agrawal, P., Fragkiadaki, K., Malik, J.: Human pose estimation
  with iterative error feedback. In: IEEE Conference on Computer Vision and
  Pattern Recognition. pp. 4733--4742 (2016)

\bibitem{du2015hierarchical}
Du, Y., Wang, W., Wang, L.: Hierarchical recurrent neural network for skeleton
  based action recognition. In: IEEE Conference on Computer Vision and Pattern
  Recognition. pp. 1110--1118 (2015)

\bibitem{duvenaud2015convolutional}
Duvenaud, D.K., Maclaurin, D., Iparraguirre, J., Bombarell, R., Hirzel, T.,
  Aspuru-Guzik, A., Adams, R.P.: Convolutional networks on graphs for learning
  molecular fingerprints. In: Advances in Neural Information Processing
  Systems. pp. 2224--2232 (2015)

\bibitem{evangelidis2014skeletal}
Evangelidis, G., Singh, G., Horaud, R.: Skeletal quads: Human action
  recognition using joint quadruples. In: IEEE International Conference on
  Pattern Recognition. pp. 4513--4518 (2014)

\bibitem{gilbert2007brain}
Gilbert, C.D., Sigman, M.: Brain states: top-down influences in sensory
  processing. Neuron  \textbf{54}(5),  677--696 (2007)

\bibitem{han2018image}
Han, W., Chang, S., Liu, D., Yu, M., Witbrock, M., Huang, T.S.: Image
  super-resolution via dual-state recurrent networks. In: IEEE Conference on
  Computer Vision and Pattern Recognition. pp. 1654--1663 (2018)

\bibitem{haris2018deep}
Haris, M., Shakhnarovich, G., Ukita, N.: Deep back-projection networks for
  super-resolution. In: IEEE Conference on Computer Vision and Pattern
  Recognition. pp. 1664--1673 (2018)

\bibitem{henaff2015deep}
Henaff, M., Bruna, J., Lecun, Y.: Deep convolutional networks on
  graph-structured data. Computer Science  (2015)

\bibitem{hu2015jointly}
Hu, J.F., Zheng, W.S., Lai, J., Zhang, J.: Jointly learning heterogeneous
  features for {RGB-D} activity recognition. In: IEEE Conference on Computer
  Vision and Pattern Recognition. pp. 5344--5352 (2015)

\bibitem{huang2019hierarchical}
Huang, L., Huang, Y., Ouyang, W., Wang, L.: Hierarchical graph convolutional
  network for skeleton-based action recognition. In: Springer International
  Conference on Image and Graphics. pp. 93--102 (2019)

\bibitem{hupe1998cortical}
Hup{\'e}, J., James, A., Payne, B., Lomber, S., Girard, P., Bullier, J.:
  Cortical feedback improves discrimination between figure and background by
  v1, v2 and v3 neurons. Nature  \textbf{394}(6695),  784--787 (1998)

\bibitem{karpathy2014large}
Karpathy, A., Toderici, G., Shetty, S., Leung, T., Sukthankar, R., Fei-Fei, L.:
  Large-scale video classification with convolutional neural networks. In: IEEE
  Conference on Computer Vision and Pattern Recognition. pp. 1725--1732 (2014)

\bibitem{ke2017new}
Ke, Q., Bennamoun, M., An, S., Sohel, F., Boussaid, F.: A new representation of
  skeleton sequences for {3D} action recognition. In: IEEE Conference on
  Computer Vision and Pattern Recognition. pp. 3288--3297 (2017)

\bibitem{ke2018learning}
Ke, Q., Bennamoun, M., An, S., Sohel, F., Boussaid, F.: Learning clip
  representations for skeleton-based {3D} action recognition. IEEE Transactions
  on Image Processing  \textbf{27}(6),  2842--2855 (2018)

\bibitem{kipf2017semi}
Kipf, T.N., Welling, M.: Semi-supervised classification with graph
  convolutional networks. In: International Conference on Learning
  Representations (2017)

\bibitem{lee1967foundations}
Lee, E.B., Markus, L.: Foundations of optimal control theory. Tech. rep.,
  Minnesota Univ Minneapolis Center For Control Sciences (1967)

\bibitem{lee2017ensemble}
Lee, I., Kim, D., Kang, S., Lee, S.: Ensemble deep learning for skeleton-based
  action recognition using temporal sliding {LSTM} networks. In: IEEE
  International Conference on Computer Vision. pp. 1012--1020 (2017)

\bibitem{li2017skeleton}
Li, B., Dai, Y., Cheng, X., Chen, H., Lin, Y., He, M.: Skeleton based action
  recognition using translation-scale invariant image mapping and multi-scale
  deep {CNN}. In: IEEE International Conference on Multimedia and Expo
  Workshops. pp. 601--604 (2017)

\bibitem{li2019actional}
Li, M., Chen, S., Chen, X., Zhang, Y., Wang, Y., Tian, Q.: Actional-structural
  graph convolutional networks for skeleton-based action recognition. In: IEEE
  Conference on Computer Vision and Pattern Recognition. pp. 3595--3603 (2019)

\bibitem{li2018independently}
Li, S., Li, W., Cook, C., Zhu, C., Gao, Y.: Independently recurrent neural
  network (indrnn): Building a longer and deeper {RNN}. In: IEEE Conference on
  Computer Vision and Pattern Recognition. pp. 5457--5466 (2018)

\bibitem{li2019learning}
Li, Y., Xia, R., Liu, X., Huang, Q.: Learning shape-motion representations from
  geometric algebra spatio-temporal model for skeleton-based action
  recognition. In: IEEE International Conference on Multimedia and Expo. pp.
  1066--1071 (2019)

\bibitem{li2019feedback}
Li, Z., Yang, J., Liu, Z., Yang, X., Jeon, G., Wu, W.: Feedback network for
  image super-resolution. In: IEEE Conference on Computer Vision and Pattern
  Recognition. pp. 3867--3876 (2019)

\bibitem{liu2017two}
Liu, H., Tu, J., Liu, M.: Two-stream {3D} convolutional neural network for
  skeleton-based action recognition. arXiv preprint arXiv:1705.08106  (2017)

\bibitem{liu2019ntu}
Liu, J., Shahroudy, A., Perez, M.L., Wang, G., Duan, L.Y., Chichung, A.K.: {NTU
  RGB+D} 120: A large-scale benchmark for {3D} human activity understanding.
  IEEE Transactions on Pattern Analysis and Machine Intelligence  (2019)

\bibitem{liu2019skeleton}
Liu, J., Shahroudy, A., Wang, G., Duan, L.Y., Chichung, A.K.: Skeleton-based
  online action prediction using scale selection network. IEEE Transactions on
  Pattern Analysis and Machine Intelligence  (2019)

\bibitem{liu2017skeleton}
Liu, J., Shahroudy, A., Xu, D., Kot, A.C., Wang, G.: Skeleton-based action
  recognition using spatio-temporal {LSTM} network with trust gates. IEEE
  Transactions on Pattern Analysis and Machine Intelligence  \textbf{40}(12),
  3007--3021 (2017)

\bibitem{liu2017skeleton-tip}
Liu, J., Wang, G., Duan, L.Y., Abdiyeva, K., Kot, A.C.: Skeleton-based human
  action recognition with global context-aware attention {LSTM} networks. IEEE
  Transactions on Image Processing  \textbf{27}(4),  1586--1599 (2017)

\bibitem{liu2017enhanced}
Liu, M., Liu, H., Chen, C.: Enhanced skeleton visualization for view invariant
  human action recognition. Elsevier Pattern Recognition  \textbf{68},
  346--362 (2017)

\bibitem{luo2013group}
Luo, J., Wang, W., Qi, H.: Group sparsity and geometry constrained dictionary
  learning for action recognition from depth maps. In: IEEE International
  Conference on Computer Vision. pp. 1809--1816 (2013)

\bibitem{niepert2016learning}
Niepert, M., Ahmed, M., Kutzkov, K.: Learning convolutional neural networks for
  graphs. In: IEEE International Conference on Machine Learning. pp. 2014--2023
  (2016)

\bibitem{ohn2013joint}
Ohn-Bar, E., Trivedi, M.: Joint angles similarities and {HOG2} for action
  recognition. In: IEEE Conference on Computer Vision and Pattern Recognition
  Workshops. pp. 465--470 (2013)

\bibitem{papadopoulos2019vertex}
Papadopoulos, K., Ghorbel, E., Aouada, D., Ottersten, B.: Vertex feature
  encoding and hierarchical temporal modeling in a spatial-temporal graph
  convolutional network for action recognition. arXiv preprint arXiv:1912.09745
   (2019)

\bibitem{parlos1994application}
Parlos, A.G., Chong, K.T., Atiya, A.F.: Application of the recurrent multilayer
  perceptron in modeling complex process dynamics. IEEE Transactions on Neural
  Networks  \textbf{5}(2),  255--266 (1994)

\bibitem{perez2019interaction}
Perez, M., Liu, J., Kot, A.C.: Interaction relational network for mutual action
  recognition. arXiv preprint arXiv:1910.04963  (2019)

\bibitem{rahmani2014real}
Rahmani, H., Mahmood, A., Huynh, D.Q., Mian, A.: Real time action recognition
  using histograms of depth gradients and random decision forests. In: IEEE
  Winter Conference on Applications of Computer Vision. pp. 626--633 (2014)

\bibitem{rahmani2015learning}
Rahmani, H., Mian, A.: Learning a non-linear knowledge transfer model for
  cross-view action recognition. In: IEEE Conference on Computer Vision and
  Pattern Recognition. pp. 2458--2466 (2015)

\bibitem{shahroudy2016ntu}
Shahroudy, A., Liu, J., Ng, T.T., Wang, G.: {NTU RGB+D}: A large scale dataset
  for {3D} human activity analysis. In: IEEE Conference on Computer Vision and
  Pattern Recognition. pp. 1010--1019 (2016)

\bibitem{shao2020learning}
Shao, Z., Li, Y., Zhang, H.: Learning representations from skeletal
  self-similarities for cross-view action recognition. IEEE Transactions on
  Circuits and Systems for Video Technology  (2020)

\bibitem{shi2019skeleton}
Shi, L., Zhang, Y., Cheng, J., Lu, H.: Skeleton-based action recognition with
  directed graph neural networks. In: IEEE Conference on Computer Vision and
  Pattern Recognition. pp. 7912--7921 (2019)

\bibitem{shi2019two}
Shi, L., Zhang, Y., Cheng, J., Lu, H.: Two-stream adaptive graph convolutional
  networks for skeleton-based action recognition. In: IEEE Conference on
  Computer Vision and Pattern Recognition. pp. 12026--12035 (2019)

\bibitem{si2019attention}
Si, C., Chen, W., Wang, W., Wang, L., Tan, T.: An attention enhanced graph
  convolutional {LSTM} network for skeleton-based action recognition. In: IEEE
  Conference on Computer Vision and Pattern Recognition. pp. 1227--1236 (2019)

\bibitem{si2018skeleton}
Si, C., Jing, Y., Wang, W., Wang, L., Tan, T.: Skeleton-based action
  recognition with spatial reasoning and temporal stack learning. In: Springer
  European Conference on Computer Vision. pp. 103--118 (2018)

\bibitem{simonyan2014two}
Simonyan, K., Zisserman, A.: Two-stream convolutional networks for action
  recognition in videos. In: Advances in Neural Information Processing Systems.
  pp. 568--576 (2014)

\bibitem{song2017end}
Song, S., Lan, C., Xing, J., Zeng, W., Liu, J.: An end-to-end spatio-temporal
  attention model for human action recognition from skeleton data. In: AAAI
  Conference on Artificial Intelligence. pp. 4263--4270 (2017)

\bibitem{stollenga2014deep}
Stollenga, M.F., Masci, J., Gomez, F., Schmidhuber, J.: Deep networks with
  internal selective attention through feedback connections. In: Advances in
  Neural Information Processing Systems. pp. 3545--3553 (2014)

\bibitem{tang2018deep}
Tang, Y., Tian, Y., Lu, J., Li, P., Zhou, J.: Deep progressive reinforcement
  learning for skeleton-based action recognition. In: IEEE Conference on
  Computer Vision and Pattern Recognition. pp. 5323--5332 (2018)

\bibitem{thakkar2018part}
Thakkar, K., Narayanan, P.: Part-based graph convolutional network for action
  recognition. In: British Machine Vision Conference. pp. 1--13 (2018)

\bibitem{toshev2014deeppose}
Toshev, A., Szegedy, C.: Deeppose: Human pose estimation via deep neural
  networks. In: IEEE Conference on Computer Vision and Pattern Recognition. pp.
  1653--1660 (2014)

\bibitem{vemulapalli2014human}
Vemulapalli, R., Arrate, F., Chellappa, R.: Human action recognition by
  representing {3D} skeletons as points in a lie group. In: IEEE Conference on
  Computer Vision and Pattern Recognition. pp. 588--595 (2014)

\bibitem{wang2013learning}
Wang, J., Liu, Z., Wu, Y., Yuan, J.: Learning actionlet ensemble for {3D} human
  action recognition. IEEE Transactions on Pattern Analysis and Machine
  Intelligence  \textbf{36}(5),  914--927 (2013)

\bibitem{wang2014cross}
Wang, J., Nie, X., Xia, Y., Wu, Y., Zhu, S.C.: Cross-view action modeling,
  learning and recognition. In: IEEE Conference on Computer Vision and Pattern
  Recognition. pp. 2649--2656 (2014)

\bibitem{yan2018spatial}
Yan, S., Xiong, Y., Lin, D.: Spatial temporal graph convolutional networks for
  skeleton-based action recognition. In: AAAI Conference on Artificial
  Intelligence. pp. 7444--7452 (2018)

\bibitem{zamir2017feedback}
Zamir, A.R., Wu, T.L., Sun, L., Shen, W.B., Shi, B.E., Malik, J., Savarese, S.:
  Feedback networks. In: IEEE Conference on Computer Vision and Pattern
  Recognition. pp. 1308--1317 (2017)

\bibitem{zhang2017view}
Zhang, P., Lan, C., Xing, J., Zeng, W., Xue, J., Zheng, N.: View adaptive
  recurrent neural networks for high performance human action recognition from
  skeleton data. In: IEEE International Conference on Computer Vision. pp.
  2117--2126 (2017)

\bibitem{zhang2012microsoft}
Zhang, Z.: Microsoft kinect sensor and its effect. IEEE Multimedia
  \textbf{19}(2),  4--10 (2012)

\bibitem{zhao2017two}
Zhao, R., Ali, H., Van~der Smagt, P.: Two-stream {RNN/CNN} for action
  recognition in {3D} videos. In: IEEE International Conference on Intelligent
  Robots and Systems. pp. 4260--4267 (2017)

\bibitem{zhao2019bayesian}
Zhao, R., Wang, K., Su, H., Ji, Q.: Bayesian graph convolution {LSTM} for
  skeleton based action recognition. In: IEEE International Conference on
  Computer Vision. pp. 6882--6892 (2019)

\end{thebibliography}
\end{document}